\documentclass[conference]{IEEEtran}
\IEEEoverridecommandlockouts
\usepackage{cite}
\usepackage{amsmath,amssymb,amsfonts}
\usepackage{algorithmic}
\usepackage{graphicx}
\usepackage{textcomp}
\usepackage{xcolor}
\usepackage{tcolorbox}  
\usepackage{amsmath}    
\usepackage{graphicx}   
\usepackage{multirow}
\usepackage{makecell}
 \usepackage{url} 
\def\BibTeX{{\rm B\kern-.05em{\sc i\kern-.025em b}\kern-.08em
    T\kern-.1667em\lower.7ex\hbox{E}\kern-.125emX}}
\begin{document}

\title{PANER: A Paraphrase-Augmented Framework for Low-Resource Named Entity Recognition
}

\author{
\IEEEauthorblockN{1\textsuperscript{st} Nanda Kumar Rengarajan}
\IEEEauthorblockA{ \textit{Concordia University}\\
nanda.kumark@mail.concordia.ca}
\and

\IEEEauthorblockN{2\textsuperscript{nd} Jun Yan}
\IEEEauthorblockA{
\textit{Concordia University}\\
jun.yan@concordia.ca}
\and

\IEEEauthorblockN{3\textsuperscript{rd} Chun Wang}
\IEEEauthorblockA{
\textit{Concordia University}\\
chun.wang@concordia.ca}
}

\maketitle

\begin{abstract}
Named Entity Recognition (NER) is a critical task that requires substantial annotated data, making it challenging in low-resource scenarios where label acquisition is expensive. While zero-shot and instruction-tuned approaches have made progress, they often fail to generalize to domain-specific entities and do not effectively utilize limited available data. We present a lightweight few-shot NER framework that addresses these challenges through two key innovations: (1) a new instruction tuning template with a simplified output format that combines principles from prior IT approaches to leverage the large context window of recent state-of-the-art LLMs; (2) introducing a strategic data augmentation technique that preserves entity information while paraphrasing the surrounding context, thereby expanding our training data without compromising semantic relationships. Experiments on benchmark datasets show that our method achieves performance comparable to state-of-the-art models on few-shot and zero-shot tasks, with our few-shot approach attaining an average F1 score of 80.1 on the CrossNER datasets. Models trained with our paraphrasing approach show consistent improvements in F1 scores of up to 17 points over baseline versions, offering a promising solution for groups with limited NER training data and compute power.
\end{abstract}


\section{Introduction}
Named Entity Recognition (NER) is a foundational task in Natural Language Processing (NLP), enabling applications like information extraction, question answering, and event detection \cite{b1}. Traditional NER systems rely on supervised learning, requiring extensive annotated data for specific domains and predefined entity types. This dependency on large, labelled datasets limits their adaptability to new domains and entity categories.
Recent breakthroughs in Large Language Models (LLMs) have enabled more flexible NER approaches through instruction tuning, demonstrating promising zero-shot and few-shot capabilities without extensive labelled data. Approaches like InstructUIE \cite{b2} and UniversalNER \cite{b3} show strong generalization across diverse entity types. However, these methods often underperform in specialized domains and face practical constraints, either requiring substantial computational resources or suffering from slow inference times \cite{b4}.

\begin{figure}[htbp]
\centerline{\includegraphics[width=0.45\textwidth]{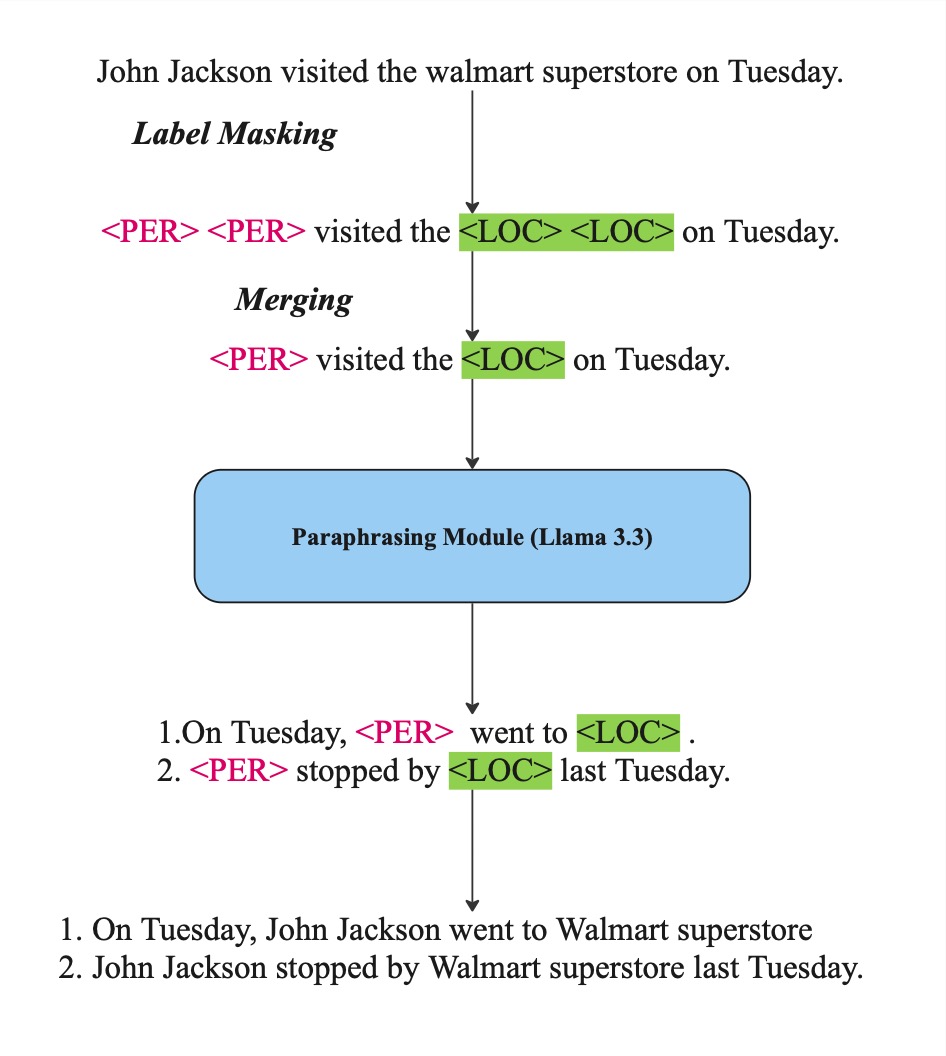}}
\caption{Illustration of paraphrasing-based data augmentation process.}
\label{fig:1}
\end{figure}
Developments in instruction tuning-based NER approaches have introduced key innovations that inspire our approach. 
The SLIMER \cite{b4} framework emphasizes the use of enriched prompts, incorporating definitions and annotation guidelines to improve performance on unseen entities. GNER \cite{b5} highlights the importance of negative instances, improving contextual understanding and entity boundary delineation by including non-entity text. We combine the strengths of both approaches—the negative instance inclusion suggested by GNER \cite{b5} and the guideline-centric philosophy of SLIMER \cite{b4} — to create an instruction-tuning template that clearly defines entity boundaries and types. Our framework adopts a simplified "word/tag" output format, reducing complexity, particularly in low-data scenarios. This integration balances robust entity representation with efficient domain adaptation. 

We also focus on the expanded context lengths offered by modern instruction-tuned models. Specifically, we leverage the 128k context windows of Qwen-2.5-Instruct (7B) \cite{b6} and LLAMA-3.1-Instruct (8B) \cite{b7}, as well as the 32k context window of Falcon3-Instruct (10B) \cite{b8}, enabling our framework to process longer and more complex inputs efficiently. These extended context lengths, combined with enriched instruction tuning, provide a foundation for robust and scalable NER performance across diverse and challenging tasks. 

Data augmentation has emerged as a key strategy for addressing limited training data in NER.  Traditional methods focus on generating augmented samples through back-translation \cite{b9}, and entity-controlled generation using question-answering techniques \cite{b10}. MELM \cite{b29} introduced a data augmentation strategy that injects entity labels into the training context, reducing token-label misalignment and improving entity diversity, particularly in low-resource and multilingual NER settings.

We introduce a targeted approach that modifies only the context surrounding entities. This preserves semantic relationships while expanding the linguistic variety, as illustrated in Figure \ref{fig:1}.  LLM-DA \cite{b28} has already demonstrated the effectiveness of large language models in generating diverse training examples by employing context-level rewriting strategies. Our approach builds on this idea by also working on sentences with multiple entities, ensuring paraphrased variants remain semantically consistent and merging entity representations to refine NER model predictions. Overall, this technique improves model adaptability to domain-specific entities and helps bridge the performance gap in low-resource settings. 

Our experiments on benchmark datasets, including CrossNER~\cite{b20} and MIT~\cite{b23}, demonstrate that our zero-shot framework achieves comparable performance to state-of-the-art zero-shot models while requiring fewer computational resources. Models using the paraphrasing augmentation consistently outperform baseline versions (without augmented data), validating the data strategy. Our key contributions include: 
\begin{itemize}
\item An instruction tuning template that combines negative instance style annotation with definition- and guideline-based schema.
\item A controlled paraphrasing technique for generating high-quality samples combined with an easily replicable low-resource training method for domain-specific NER. 
\end{itemize}
The remainder of this paper is organized as follows. {Section \ref{sec:2}} reviews related work in Named Entity Recognition (NER), with a focus on zero-shot and few-shot learning using Large Language Models (LLMs). {Section \ref{sec:3}} describes the proposed methodology, detailing the paraphrasing-based data augmentation framework and the implementation details, including optimization strategies and validation techniques. {Section \ref{sec:4}} talks about the instruction tuning approach and template-used. {Section \ref{sec:5}} presents the experimental setup, covering dataset selection, baseline models, and evaluation metrics. {Section \ref{sec:6}} reports the experimental results and provides a comparative analysis of model performance. {Section \ref{sec:7}} discusses key findings, limitations, and implications for future research. 

\section{Related Work}\label{sec:2}
Named Entity Recognition (NER) has traditionally been approached as a sequence labelling task, where models are trained to assign BIO (Beginning, Inside, Outside) tags to input tokens \cite{b1}. While supervised approaches using BERT-based architectures have shown strong performance, they remain constrained by their reliance on predetermined label sets and domain-specific training data \cite{b11}.

The emergence of Large Language Models (LLMs) has introduced new paradigms for addressing NER challenges, particularly in zero-shot and few-shot scenarios. Early work by \cite{b12} demonstrated LLMs' capacity for multi-task learning through natural language instructions, laying the groundwork for what would become known as in-context learning and prompt engineering. However, initial attempts to apply LLMs to Information Extraction tasks, including NER, revealed significant limitations compared to traditional supervised approaches \cite{b2}, \cite{b13}.

Few-shot NER remains challenging due to knowledge transfer limitations across domains. Recent research has explored various instruction-tuning strategies to enhance LLMs' performance on NER tasks. Notable approaches include InstructUIE \cite{b2}, which utilizes a T5-11B architecture fine-tuned on information extraction datasets, and UniNER \cite{b3}, which employs a conversational template with LLAMA. These methods have shown promise but often require substantial computational resources for training and inference. GoLLIE \cite{b14} introduced an innovative approach by incorporating annotation guidelines such as Python docstrings, marking the first attempt to encode labelling criteria within the prompt structure explicitly.
While instruction tuning has emerged as a leading method for generalization to unseen tasks \cite{b15}, \cite{b16} current approaches face several limitations. 
\begin{enumerate}
    \item Many methods rely solely on label names in prompts without considering domain-specific definitions or complex label semantics \cite{b17}. 
    \item The computational requirements of these models often make them impractical for organizations with limited resources. Additionally, the challenge of effectively utilizing available domain data remains largely unaddressed in current instruction-tuning frameworks.
\end{enumerate}

\begin{figure}[h!]
\centering
\begin{tcolorbox}[colframe=black, colback=gray!10, width=\columnwidth]
\textbf{Paraphrasing Prompt}

\textbf{Task Description:}\\
You are a helpful assistant. I have a sentence with certain entities that I want to preserve in spirit, but you may modify the sentence slightly to add variety. Your task is:
\begin{enumerate}
	\item Read the Original Sentence provided.
    \item Create 2 new sentences (variants) that:
    \begin{itemize}
        \item DO NOT MODIFY any word enclosed in \textless\textless\textgreater\textgreater \  tags or move them around (do not introduce any new \textless\textless\textgreater\textgreater \ tags that weren't in the original).
        \item May adjust phrasing, structure, or add contextual details while maintaining logical coherence and meaning.
        \item Minor modifications are allowed, but retain the core entity references and do not transform them into something else.
    \end{itemize}
	\item Return the output in a valid JSON format with the generated variants.
\end{enumerate}
\textbf{Original Sentence}: {\textit{Input}}

\end{tcolorbox}

\caption{Prompt used for generating paraphrases.}
\label{fig:para_prompt}
\end{figure}

Recent studies have begun to address the efficiency-performance trade-off in NER systems. GNER \cite{b5} introduced a modified BIO-like generation approach that improves boundary detection while addressing classification indecision. GLiNER \cite{b18} demonstrated that smaller, non-instruction-tuned architectures could achieve competitive performance in both supervised and zero-shot settings, suggesting the potential for more resource-efficient approaches.

\section{Proposed Method}\label{sec:3}
Our data augmentation step, illustrated in Figure \ref{fig:1}, was developed to address a fundamental challenge in Named Entity Recognition (NER) - the scarcity of annotated training data. We leveraged the LLAMA 3.3-70B \cite{b24} model for generating paraphrases, using the prompt shown in Fig. \ref{fig:para_prompt}. We chose this model for its ability to match the performance of larger models like LLAMA 3.1-405B while operating with substantially fewer parameters (70B vs 405B), making it more practical for real-world applications \cite{b19}.

\subsection{Paraphrasing Framework}\label{11}
The core of our approach involves transforming input sentences into masked templates where named entities are replaced with semantic placeholders. For example, given the input sentence ``John visited the supermarket on Tuesday,'' the system generates a masked version: ``\textless PER\textgreater visited the \textless LOC\textgreater   on Tuesday.'' This masking preserves the essential structure of the sentence while marking entity positions for consistent paraphrasing.
Using these masked templates, the LLAMA 3.3-70B model generates variations that maintain the original entity relationships while introducing some diversity. Our experiments produced paraphrases such as:
\begin{itemize}
    \item \textless PER\textgreater\ stopped by \textless LOC\textgreater\ last Tuesday
    \item On Tuesday, \textless PER\textgreater\ went to \textless LOC\textgreater
    \item Tuesday saw \textless PER\textgreater\ traveling to \textless LOC\textgreater
\end{itemize}

Through experimental evaluation, we determined that generating two paraphrased versions per input sentence achieved the optimal balance between diversity and quality. Attempts to produce three or more variations frequently resulted in redundancy or significant deviations from the original meaning. Furthermore, the additional paraphrases often failed to adhere to the specified output formatting requirements (JSON format) and occasionally produced sentences with an inconsistent number of \textless ENT\textgreater tags compared to the input. Based on these observations, we limited the generation to two variants. However, the system remains configurable, allowing users to generate additional variations by adjusting the temperature parameter during generation.

\subsection{Implementation and Optimization}
We have iterated over the paraphrasing prompt multiple times and included optimizations to enhance the reliability of the proposed paraphrasing system. One key improvement involved the handling of consecutive entity tags. When multiple words belonging to the same entity type appear in sequence (for example, a four-word organization name), we consolidate them into a single entity tag rather than using multiple consecutive tags. This simplification reduced the complexity of the paraphrasing task and improved the model's ability to maintain entity consistency.

Initially, we used generic \textless ENTITY\textgreater tags instead of specific tags like \textless PER\textgreater or \textless LOC\textgreater. However, this approach proved less effective as the model lacked sufficient context about the type of entity being masked. Specific entity tags provided better guidance for the paraphrasing process, particularly in domain-specific contexts where entity relationships are more nuanced.

\subsection{Quality Control and Validation}
To ensure the quality of generated paraphrases, a structured validation pipeline was developed using the instructor package and a locally hosted version of LLAMA. Our system processes the model's output in JSON format, allowing efficient parsing and validation of the generated paraphrases. For each paraphrase, we verify that:
\begin{enumerate}
    \item The number of entity tags matches the input sentences.
    \item The semantic relationships between entities are preserved (cosine similarity).
\end{enumerate}

When a generated paraphrase fails these validation checks, the system either triggers a regeneration with adjusted parameters or attempts to map the entities correctly based on their position and context in the original sentence. This validation process helps maintain the integrity of the augmented dataset while allowing for natural variations in sentence structure and word choice.
The complete paraphrasing prompt is shown in Fig. \ref{fig:para_prompt}, where we instruct the model to maintain entity references while allowing for structural variations and additional context.

\begin{figure}[h!]
\centering
\begin{tcolorbox}[colframe=black, colback=gray!10, width=\columnwidth]
\textbf{Instruction Tuning Prompt}

\textbf{Task Description:}\\
Please analyze the sentence provided, identifying the type of entity for each word on a token-by-token basis.
Each word in the sentence should be annotated with its corresponding named entity tag, using a forward slash / between the word and the tag. Output format is: \texttt{word\_1/label\_1, word\_2/label\_2, ...}

\textbf{Guideline:}
\begin{enumerate}
    \item Use O for words that are not part of any named entity. 
    \item For multi-word entities, label each word with the same entity tag.
\end{enumerate}

Use the specific entity tags: $l_1, l_2, \ldots, l_m$, and \textbf{O}.
To help you, here are dedicated DEFINITION and GUIDELINES for each entity tag. \\
\texttt{\{ $l_1$ : \{ \\ 
DEFINITION : , \\
GUIDELINES :  \} \\ 
\}} \\ 
\textbf{Input:} $x_1 \quad x_2 \quad \ldots \quad x_n$

\textbf{Output:} $x_1/\hat{y}_1 \quad x_2\hat{y}_2 \quad \ldots \quad x_n/\hat{y}_n$
\end{tcolorbox}

\caption{Prompt used for Instruction-tuning LLMs.}
\label{fig:prompt}
\end{figure}

\section{Instruction Tuning and Adaptation of Prompt Design}\label{sec:4}

In this work, we revisit and refine the instruction-tuning methodologies outlined in GNER \cite{b5}, diverging from the traditional BIO tagging schema in favour of a word/tag representation format. The proposed method annotates each word with its corresponding entity tag using a forward slash (/) separator, simplifying the tagging process by removing the complexity of distinguishing between ``B-'' and ``I-'' labels. Notably, all words within multi-word entities are assigned the same tag. Furthermore, we provide detailed definitions and guidelines, the same way as SLIMER \cite{b4}, for each entity type to enhance the extraction.

We continue to leverage the effective strategy of incorporating negative instances, as outlined by GNER \cite{b5}. This technique helps the model differentiate between entity and non-entity tokens by utilizing surrounding context, reducing reliance on direct memorization of entity names. By preserving this contextual learning mechanism and combining it with our simplified tagging format, we establish a robust framework that enhances performance for entity recognition tasks. Table \ref{tab:format_comp} compares the result of different formats for instruction tuning. 

When applying this new instruction prompt to the CrossNER \cite{b20} Science dataset with 16 entity types in the training prompt, including task description, annotations, and guidelines for all NEs, added up to 1700 tokens, well below the context length of the models used in our experiments. This token efficiency allowed us to include comprehensive task instructions and entity definitions directly in the input prompts without exceeding the model’s context limit. The results demonstrate that this format is not only feasible but also beneficial for few-shot and zero-shot settings.

We employ LoRA \cite{b21} fine-tuning for a single epoch to further optimize training, reducing computational and memory requirements while achieving performance comparable to full fine-tuning methods. The goal is to achieve comparable or superior performance using a fraction of the resources, particularly for low-resource NER datasets. 

For datasets with longer sentences or complex annotations, a chunking strategy is used to split inputs into manageable segments while preserving overall context. To ensure reliable response generation, the context length is limited to 2048 tokens. As a result, any sequences exceeding this threshold are automatically segmented into multiple examples. Fig. \ref{fig:prompt} shows the full instruction tuning prompt used. 

\section{Experiment Setup}\label{sec:5}

\subsection{Datasets}\label{datasets}

We use PileNER \cite{b3} as the main training corpus, with key pre-processing steps to ensure data quality and consistency. The pre-processing pipeline includes the following filtering criteria:
\begin{enumerate}
    \item Minimum sentence length threshold of 10 words to ensure sufficient context.
    \item Language filtering to retain only English text.
    \item Entity type filtering to focus on 423 named entities with established guidelines and annotations, as documented by \cite{b4}
\end{enumerate}

This filtered dataset yielded approximately 23,402 high-quality samples. To maintain experimental consistency with prior work \cite{b5}, 10,000 samples are randomly selected from this preprocessed pool as a starting point for our few-shot testing. 

We evaluate the proposed approach on four established benchmarks, each chosen to assess different aspects of model performance:
\begin{enumerate}
    \item CrossNER \cite{b20}: A comprehensive cross-domain dataset that evaluates domain adaptation capabilities across diverse subject areas, including scientific papers, politics, music, and literature.
    \item MIT \cite{b23}: A standard benchmark for assessing out-of-distribution (OOD) performance, particularly valuable for evaluating generalization to novel domains.
    \item  BUSTER \cite{b22}: A document-level financial domain NER benchmark that presents unique challenges through its specialized entity types and complex document structure.
    \item CoNLL \cite{b35}: CoNLL-2003 shared task dataset is a widely-used benchmark for NER featuring multilingual annotated text with general tags like PERSON, LOCATION, and others. 
\end{enumerate}

\subsection{Baseline Comparisons}
To evaluate the effectiveness of our approach, we compare it against several state-of-the-art methods for zero-shot and few-shot Named Entity Recognition (NER) and data augmentation. Each baseline represents a distinct methodology or model architecture, providing a diverse comparison framework.
\begin{enumerate}
        \item GoLLIE \cite{b14}: A generative model based on Code-LLAMA, designed to leverage annotation guidelines formatted in a code-like representation. We use the 7B variant of GoLLIE for comparability with other models in this study.
	\item GLiNER-L \cite{b18}: An encoder-only model based on DeBERTa with 304 million parameters. Despite being the smallest model among the selected baselines, GLiNER-L has demonstrated competitive performance in out-of-distribution (OOD) zero-shot NER tasks.
	\item GNER \cite{b5}: A model released in two variants, each leveraging a different backbone architecture:
    \begin{itemize}
        \item \textit{GNER-T5}: Based on flan-t5-xxl.
        \item \textit{GNER-LLAMA}: Built on the LLAMA-7B architecture. Both versions emphasize the incorporation of entity definitions during instruction-tuning.
    \end{itemize}
    \item SLIMER \cite{b4}: A model based on the LLAMA-2-7B chat architecture, fine-tuned with LoRA \cite{b21} for 10 epochs. SLIMER integrates structured annotation guidelines, making it a strong benchmark for guideline-based NER.
    \item DAGA \cite{b34}:  DAGA utilizes a one-layer LSTM-based language model trained on linearized labelled sentences from CoNLL and other sequence-tagging datasets to generate synthetic training data for the same.
    \item MELM \cite{b29}: a data augmentation framework that ensures label-consistent entity replacements by fine-tuning XLM-RoBERTa with masked entity prediction.
\end{enumerate}

\subsection{Backbone LLMs and Evaluation Framework}\label{sec:backbone}
We used Qwen-2.5-Instruct (7B), LLAMA-3.1-Instruct (8B), and Falcon3-Instruct (10B) as our backbone models, selected based on their extended context lengths: 128K for Qwen-2.5 and LLAMA-3.1, and 32K for Falcon3, respectively, as well as their state-of-the-art performance on instruction-following benchmarks such as MT-Bench \cite{b25} and Alpaca WC \cite{b26}. 

Our evaluation strategy includes both few-shot and zero-shot scenarios to assess model performance under varying resource constraints. For zero-shot evaluation, we fine-tuned the models above on 23,402 pileNER samples, as described in Section \ref{datasets}. Although we utilize more samples than \cite{b5}, our training setup (described below) is significantly more efficient and effectively bridges the performance gap while still serving as a cost-effective solution.
For few-shot evaluation, 
we used 10,000 samples from pileNER as our base dataset and added domain-specific examples from the benchmark datasets (CrossNER and MIT) as necessary.  


\textbf{Training Setup: } All models were fine-tuned on the Modal platform using Axolotl. Fine-tuning was conducted with LoRA~\cite{b21} settings of $r = 8$, $\alpha = 16$, and the AdamW optimizer~\cite{b33} for one epoch. A cosine learning rate schedule was employed, starting with a warm-up phase covering 4\% of the training steps and peaking at $2 \times 10^{-5}$. 

\begin{table}[]
\caption{Comparison between Instruction Formats}
\begin{tabular}{lcccccc}
\hline
\textbf{} & \textbf{AI} & \textbf{Lit} & \textbf{Music} & \textbf{Pol} & \textbf{Sci} & \textbf{Avg} \\ \hline
GNER-BIO & 52.1 & 51.1 & 58.5  & 54.1 & 43.8 & 51.92 \\
Ours-slash w/o* & 59.1 & 67.4 & 72.25 & 70.8 & 66.1 & 67.13 \\
Ours-slash & 63.9 & 67.2 & 75.3 & 67.8 & 68.7 & \textbf{68.58} \\ \hline
\label{tab:format_comp}
\end{tabular}
 \\ *w/o: prompt without guidelines
\end{table}

\begin{table*}[!t]
\renewcommand{\arraystretch}{1.3}
\caption{Few-shot F1 (\%) scores using augmented samples Across Different Domains}
\centering
\resizebox{\textwidth}{!}{ 
\begin{tabular}{|l|c|c|c|c|c|c|c|c|c|c|}
\hline
\textbf{Model Family} & \makecell{\textbf{\# of} \\ \textbf{Original} \\ \textbf{Samples}} & \makecell{\textbf{\# of} \\ \textbf{Augmented} \\ \textbf{Samples}} & \textbf{Movie} & \textbf{Restaurant} & \textbf{AI} & \textbf{Literature} & \textbf{Music} & \textbf{Politics} & \textbf{Science} & \textbf{Average} \\ \hline
LLAMA-3.1-8B-Instruct & 0 & 0 & 43.3 & 30.3 & 59.4 & 61.5 & 68.2 & 62.1 & 62.3 & 55.3 \\ \hline
\multirow{2}{*}{ } & 100 & 0 & 45.1 & 35.4 & 61.0 & 67.2 & 75.9 & 70.4 & 70.9 & 60.8 \\ \cline{2-11}
 & 100 & 200 & 64.2 & 39.8 & 64.0 & 77.0 & 80.8 & 73.2 & 72.9 & \textbf{67.4} \\ \hline
Qwen-2.5-7B-Instruct & 0 & 0 & 50.3 & 33.6 & 46.5 & 54.9 & 53.9 & 54.0 & 49.1 & 48.9 \\ \hline
\multirow{2}{*}{ } & 100 & 0 & 54.2 & 25.6 & 57.8 & 63.4 & 70.4 & 64.2 & 65.9 & 57.4 \\ \cline{2-11}
 & 100 & 200 & 65.1 & 38.2 & 59.5 & 73.3 & 79.2 & 70.2 & 75.3 & \textbf{65.8} \\ \hline
Falcon-3-10B-Instruct  & 0 & 0 & 64.5 & 38.1 & 63.7 & 58.8 & 68.6 & 61.8 & 59.4 & 59.3 \\ \hline
\multirow{2}{*}{ }  & 100 & 0 & 63.7 & 37.0 & 67.8 & 67.6 & 82.2 & 72.0 & 77.5 & 66.8 \\ \cline{2-11}
  & 100 & 200 & 77.5 & 42.8 & 72.7 & 79.0 & 85.3 & 81.3 & 82.3 & \textbf{74.4} \\ \hline
\end{tabular}
}
\label{tab:few_shot_self}
\end{table*}

\begin{table}[htbp]
\caption{Comparison of F1 (\%) scores on CrossNER for supervised techniques}
\begin{tabular}{lcccccc}
\hline
\textbf{} & \textbf{AI} & \textbf{Lit} & \textbf{Music} & \textbf{Pol} & \textbf{Sci} & \textbf{Avg} \\ \hline
BERT & 68.7 & 64.9 & 68.3 & 63.6 & 58.8 & 64.9 \\
CDLM & 68.4 & 64.3 & 63.5 & 59.5 & 53.7 & 61.9\\
DAPT & 72.0 & 68.8 & 75.7 & 69.0 & 62.6 & 69.6\\ 
NER-BERT & \textbf{76.1} & 72.1 & 80.2 & 71.9 & 63.3 & 72.7\\ 
\textbf{PANER (\textit{Ours}}) & 72.7 & \textbf{79} & \textbf{85.3} & \textbf{81.3} & \textbf{82.3} & \textbf{80.1}\\ \hline
\label{tab:crossner_supervised}
\end{tabular}
\end{table}

\begin{table}[htbp]
\caption{Comparison of F1 (\%) scores on CrossNER for augmentation composition with Llama - 3.1-8B-instruct}
\begin{tabular}{lcccccc}
\hline
\textbf{} & \textbf{AI} & \textbf{Lit} & \textbf{Music} & \textbf{Pol} & \textbf{Sci} & \textbf{Avg} \\ \hline
100 OG + 200 dup & 60.8 & 67.5 & 72.7 & 68.4 & 64.9 & 66.8\\
100 OG + 200 aug & 63.9 & 77.1 & 80.2 & 71.9 & 72.7 & 73.2\\ 
300 OG & 67.4 & 79.0 & 80.1 & 73.3 & 76.5 & 75.3\\ 
\hline
\label{tab:crossner_augmented}
\end{tabular}
\end{table}

\begin{table}[ht]
    \centering
    \caption{Performance comparison of different augmentation methods on English (En)}
    \begin{tabular}{|c |l |c|}
        \hline
        \textbf{\#Gold} & \textbf{Method} & \textbf{F1 Score (in \%)} \\
        \hline
        \multirow{6}{*}{100} & Gold-Only & 50.57 \\
        & Label-wise & 61.34 \\
        & MLM-Entity & 61.22 \\
        & DAGA & 68.06 \\
        & MELM & \textbf{75.21} \\
        & \textbf{PANER (Ours)} & \textbf{80.52} \\
        \hline
        \multirow{6}{*}{200} & Gold-Only & 74.64 \\
        & Label-wise & 76.82 \\
        & MLM-Entity & 79.16 \\
        & DAGA & 79.11 \\
        & MELM & \textbf{82.91} \\
        & \textbf{PANER (Ours)} & \textbf{85.74} \\
        \hline
        \multirow{6}{*}{400} & Gold-Only & 81.85 \\
        & Label-wise & 84.62 \\
        & MLM-Entity & 83.82 \\
        & DAGA & 84.36 \\
        & MELM & \textbf{85.73} \\
        & \textbf{PANER (Ours)} & \textbf{88.11} \\
        \hline
    \end{tabular}
    \label{tab:augment_results}
\end{table}

\begin{table*}[!t]
\renewcommand{\arraystretch}{1.3}
\caption{Comparison of Zero-shot Learning Performance F1 (\%) scores}
\centering
\resizebox{\textwidth}{!}{ 
\begin{tabular}{|l|c|c|c|c|c|c|c|c|c|c|}
\hline
\textbf{Model} & \textbf{Backbone} & \textbf{\#Params} & \textbf{Movie} & \textbf{Restaurant} & \textbf{AI} & \textbf{Literature} & \textbf{Music} & \textbf{Politics} & \textbf{Science} & \textbf{Average} \\ \hline
ChatGPT & gpt-3.5-turbo & - & 5.3 & 32.8 & 52.4 & 39.8 & 66.6 & 68.5 & 67 & 47.5 \\ 
InstructUIE & Flan-T5-xxl & 11B & 63 & 21 & 49 & 47.2 & 53.2 & 48.2 & 49.3 & 47.3  \\ 
UniNER-type+sup. & LLAMA-1 & 7B & 61.2 & 35.2 & 62.9 & 64.9 & 70.6 & 66.9 & 70.8 & 61.8 \\ 
GoLLIE & Code-LLAMA & 7B & 63 & 43.4 & 59.1 & 62.7 & 67.8 & 57.2 & 55.5 & 58.4 \\ 
GLiNER-L & DeBERTa-v3 & 0.3B & 57.2 & 42.9 & 57.2 & 64.4 & 69.6 & 72.6 & 62.6 & 60.9 \\ 
GNER-T5 & Flan-T5-xxl & 11B & 62.5 & 51 & 68.2 & 68.7 & 81.2 & 75.1 & 76.7 & 69.1 \\ 
GNER-LLAMA & LLAMA-1 & 7B & 68.6 & 47.5 & 63.1 & 68.2 & 75.7 & 69.4 & 69.9 & 66.1 \\ 
SLIMER & LLAMA-2-chat & 7B & 50.9 & 38.2 & 50.1 & 58.7 & 60 & 63.9 & 56.3 & 54 \\ \hline
\textbf{PANER} & Qwen-2.5-Instruct & 7B & 51.5 & 37.3 & 62 & 61.7 & 75.9 & 69.72 & 65.63 & 60.5 \\ 
\textbf{PANER} & LLAMA-3.1-Instruct & 8B & 52 & 37 & 63.9 & 67.2 & 75.3 & 67.8 & 68.7 & 61.7 \\ 
\textbf{PANER} & Falcon3-Instruct & 10B & 69.4 & 43.3 & 65.5 & 61.3 & 75.8 & 70.3 & 68.3 & 64.8 \\ \hline
\end{tabular}
}
\label{tab:zero_shot}
\end{table*}

\begin{table*}[!t]
\renewcommand{\arraystretch}{1.3}
\caption{Zero-shot result comparison on BUSTER dataset F1 (\%) scores}
\centering
\begin{tabular}{|l|c|c|c|c|c|c|c|c|c|c|}
\hline
\textbf{Model} & \textbf{Backbone} & \textbf{\#Params} & \textbf{Pr.} & \textbf{R} & \textbf{F1} \\ \hline
GNER-LLAMA & LLAMA-1 & 7B & 14.68 & 59.97 & 23.58 \\
GLINER-L & DeBERTa-v3 & 0.3B & 42.55 & 19.31 & 26.57 \\
GoLLIE & Code-LLAMA & 7B & 28.82 & 26.63 & 27.68 \\
GNER-T5 & Flan-T5-xxl & 11B & 19.31 & 50.15 & 27.88 \\
UniNER-type+sup. & LLAMA-1 & 7B & 31.4 & 47.53 & 37.82 \\
SLIMER  & LLAMA-2-chat & 7B & 47.69 & 43.09 & \textbf{45.27} \\\hline
\textbf{PANER} & Falcon3-Instruct & 10B & 29.92 & 38.38 & \textbf{33.63} \\ \hline
\end{tabular}
\label{tab:buster}
\end{table*}

\begin{figure}[t]
\centerline{\includegraphics[width=0.49\textwidth]{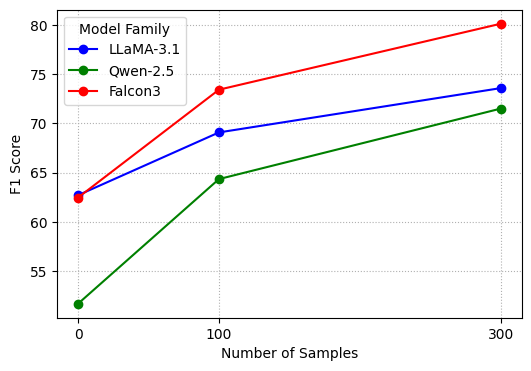}}
\caption{Impact of augmented sample size on model performance (F1 score, in \%) for CrossNER dataset~\cite{b20}.}
\label{fig:fig4}
\end{figure}

\section{RESULTS}\label{sec:6}

\subsection{Comparison of Tagging Formats} 

We first validate the effectiveness of our word/tag format by comparing it against the BIO-style output format presented in \cite{b5}. For this experiment, LLAMA 3.1-8B-Instruct was used as the backbone architecture, and we leveraged the same PileNER dataset \cite{b3} and filtered for sentences with more than 10 words only in English, resulting in approximately 23,402 samples for training. The model was fine-tuned with LoRA with the hyperparameters in Section  \ref{sec:backbone}. To maintain consistency with GNER \cite{b5}, we evaluated the model on the same five datasets, with 200 random samples per dataset, for zero-shot performance analysis. Results reported in Table \ref{tab:format_comp} represent averages over three test runs to ensure robustness. 

Reference \cite{b5} performed a similar boundary analysis with different tagging formats to determine the optimal approach. While their results differ from ours, this discrepancy could be attributed to our use of a larger model, the LLAMA 3.1-8B, compared to their Flan-T5 large model with 780M parameters. Additionally, we employ a LoRA fine-tuning approach for a single epoch, in contrast to their full fine-tuning over three epochs, which may also contribute to the observed differences in performance.

Our evaluation methodology for comparing different tagging formats, we employ entity-level F1 scores for both the BIO and word/tag formats to ensure fair comparison. While the formats differ in presentation, our evaluation criteria remain consistent across both approaches: \textit{an entity prediction is considered correct only when both the entity type and its complete boundaries match the gold standard annotation.} This means that in both formats, partial entity identification or incorrect boundary detection is treated as an error, regardless of whether the entity type was correctly identified.

Table \ref{tab:format_comp} shows a significant improvement in the F1 score when adopting the word/tag format compared to the traditional BIO tagging schema. A slight increase in F1 score is observed when definitions and guidelines are included alongside the new format. While this increase may appear marginal, it offers substantial benefits when integrated with our paraphrase-augmented synthetic data during few-shot testing (as shown in Table \ref{tab:format_comp}). This improvement underscores the complementary nature of clear guidelines and the word/tag format in enhancing model accuracy and adaptability.

\subsection{Performance of Paraphrasing in Few-shot NER}

We further evaluate the effectiveness of the paraphrasing-based data augmentation approach across multiple domain-specific datasets, including CrossNER~\cite{b20} and MIT \cite{b23}. The three models listed in Section \ref{sec:backbone} were fine-tuned using 0, 100, and 300 augmented samples with a fixed set of 10,000 PileNER \cite{b3} samples. This dataset size was chosen to align with the GNER\cite{b5} framework.

As illustrated in Figure \ref{fig:fig4}, there is an increase in F1 scores correlating with the increase in augmented samples. The results of this experiment are presented in Table \ref{tab:few_shot_self}, where we observe that the average F1 score of Falcon 3 \cite{b8} increased by 0.14, Qwen 2.5 \cite{b6} by 0.17 and LLAMA 3 \cite{b7} by 0.12, respectively. The average performance across the CrossNER \cite{b20} datasets is illustrated in Figure \ref{fig:fig4}. This indicates that the diversity introduced through the paraphrasing strategy effectively contributes to model performance.

In particular, the result using Falcon-3 \cite{b8} with 300 augmented samples achieved superior performance compared to previously reported supervised techniques on the CrossNER \cite{b20} datasets, further reinforcing our augmentation strategy. Although many recent IT studies that report on this dataset focus on zero-shot performance metrics (reported below), the results surpass few-shot and fine-tuned in which CrossNER \cite{b20}, which utilized supervised training samples. These comparative results, detailed in Table \ref{tab:crossner_supervised}, demonstrate that our data augmentation technique can enhance model performance for cases with few examples. Although the performance of other Instruction-tuned models with similar augmentation strategies remains unexplored, the results suggest that combining lightweight few-shot learning with intelligent data augmentation offers a promising direction for domain-specific NER tasks.

Additionally, we compared our paraphrasing-based data augmentation approach against existing paraphrasing techniques, including DAGA \cite{b34} and MELM \cite{b29}, on the CoNLL \cite{b35} shared task dataset. Our results indicate that leveraging LLMs for paraphrasing yields superior performance compared to these established techniques, as shown in Table \ref{tab:augment_results}.

For this experiment, we simulated a low-resource scenario for the CoNLL dataset by using 100, 200, and 400 gold samples, following the setup of \cite{b29}, and then generating 200, 400, and 800 augmented samples, respectively. We then trained the LLAMA 3.1-8B model using these configurations. Our approach consistently outperforms MELM, which utilizes 3× the number of samples compared to the 2× used in our method. This performance gain can be attributed to our use of LLMs for prediction, as they are already familiar with the entity types in the CoNLL dataset (PERSON, LOCATION, ORGANIZATION).

\subsection{Effectiveness of Paraphrase-Based Augmentation Compared to Data Duplication and In-Domain Expansion}

In order to understand the isolated effects of our paraphrasing method against duplication. We tested the comparative effectiveness of our data augmentation approach versus simply adding more in-domain samples or duplicating existing data, we conducted an additional experiment to isolate the impact of our paraphrasing-based augmentation strategy. This experiment systematically compared three training configurations, each with a total of 300 samples but differing in composition: 
\begin{enumerate}
    \item 100 original in-domain samples augmented with 200 paraphrased variants,
    \item 300 distinct original in-domain samples, and
    \item 100 original in-domain samples duplicated two times.
\end{enumerate}

The results, presented in Table~\ref{tab:crossner_augmented}, demonstrate several key findings. The configuration using 300 distinct original samples achieved the highest average F1 score (75.3\%), which was expected given the inherent value of diverse, authentic samples. However, our hybrid approach combining 100 original samples with 200 paraphrased variants performed remarkably well, reaching an F1 score of (73.2\%), only 2.1 percentage points below the all-original configuration. This suggests that our paraphrasing strategy successfully preserves the essential entity relationships while introducing beneficial linguistic variation. 

In contrast, the simple duplication approach yielded substantially lower performance (66.8\%), confirming that mere repetition of training examples provides no meaningful diversity to enhance model generalization. These findings validate our augmentation approach as an effective strategy when additional authentic in-domain samples are unavailable or prohibitively expensive to obtain, offering nearly comparable performance to training with three times the amount of original data.

Overall, these results demonstrate the effectiveness of LLM-generated paraphrases in enhancing model generalization for NER tasks, further validating our approach as a viable alternative to conventional data augmentation strategies.

\subsection{Performance of Instruction Tuning Template in Zero-shot NER}

While our primary goal is to improve few-shot Named Entity Recognition (NER), as demonstrated above, the proposed instruction-tuning template also performs competitively with state-of-the-art methods in zero-shot NER. Table \ref{tab:zero_shot} presents the zero-shot performance compared to existing state-of-the-art benchmarks. Notably, the Falcon-3 \cite{b8} model achieves an average F1 score of 0.648, close to the GNER-T5 \cite{b5} and GNER-LLAMA \cite{b5} models. This performance was obtained via fine-tuning with LoRA \cite{b21}, requiring significantly less computational time and resources: only one epoch was sufficient to achieve this performance, whereas GNER was fully fine-tuned for three epochs.

Further, the out-of-domain performance of our instruction-tuning template is showcased in Table \ref{tab:buster}, where we compare it against the above-mentioned models on the BUSTER dataset. The results demonstrate that the proposed model performs better than both GNER \cite{b5} models with an F1 of 0.336 and achieves performance comparable to SLIMER \cite{b4}, which holds the state-of-the-art F1 score of 0.4527.

\section{Conclusions}\label{sec:7}

In this work, we presented PANER, a paraphrase-augmented framework designed to enhance Named Entity Recognition (NER) in low-resource settings. The approach integrates instruction tuning with paraphrase-based data augmentation, enabling improved performance while maintaining computational efficiency. Experimental results demonstrate that PANER achieves competitive performance with state-of-the-art zero-shot NER models while requiring significantly fewer computational resources. The paraphrasing technique consistently improves entity recognition, particularly in domain-specific and few-shot learning settings. The results indicate that our approach is an effective alternative for organizations with limited access to annotated datasets and compute power.

However, our study has certain limitations. First, while paraphrasing improves model generalization, the quality of generated variations can vary depending on the complexity of the input sentences. Additionally, the approach to include guidelines and annotations for all entity types does not benefit cases where the entity is negatively affected by the guidelines. Prior work SLIMER \cite{b4} did an entity-by-entity analysis to see how guidelines and annotations are helping each entity, and the results show some entities do not require or benefit from the presence of guidelines and annotations. Since we process entire sentences and extract all entities in a single request, it is difficult to selectively include or exclude guidelines based on specific entity types, which could impact performance for certain categories. 

While our paraphrasing-based augmentation approach demonstrates promising results, we acknowledge a few limitations. First, our strict entity preservation constraints, though effective for maintaining semantic relationships, may restrict the diversity of the generated samples. During our analysis, we observed that augmented sentences often exhibit limited structural variation when multiple entities appear in close proximity, as the model prioritizes preserving entity positions over introducing novel sentence constructions. This trade-off between entity integrity and linguistic diversity represents an inherent tension in our current implementation. Additionally, the paraphrasing approach occasionally struggles with domain-specific terminology and complex syntactic structures, resulting in approximately 15\% of initially generated paraphrases failing our validation checks and requiring regeneration.

\section{Future Work}\label{sec:8}
Future work will explore more flexible entity augmentation strategies that preserve semantic relationships while allowing controlled entity variations (such as replacing entities with semantically equivalent alternatives within the same type) and adaptive paraphrasing approaches that adjust constraint strictness based on sentence complexity and domain characteristics. We also plan to investigate multi-stage augmentation pipelines that combine paraphrasing with other techniques to further enhance sample diversity while maintaining the crucial entity relationships that drive NER performance.

Another key direction for future work is refining selective guideline inclusion, where entity-specific constraints could be dynamically applied during instruction tuning. Further, while our approach has demonstrated strong performance in English-language datasets, its multilingual effectiveness remains unexplored. A critical next step is studying how paraphrase-based augmentation can be effectively applied to other languages. These advancements will ensure PANER remains a scalable, adaptable, and an efficient framework for real-world, multilingual applications.

\end{document}